\documentclass[sigplan]{acmart}
\renewcommand\footnotetextcopyrightpermission[1]{}
\usepackage{graphicx}
\usepackage{textcomp}
\usepackage{xcolor}
\usepackage{url}
\usepackage{color}
\usepackage{float}
\usepackage{multirow}
\usepackage{array}
\usepackage{todonotes}
\usepackage{comment}
\usepackage{amsmath}

\usepackage{stfloats}
\usepackage{booktabs}
\usepackage{algorithm}
\usepackage{algorithmic}
\usepackage{upgreek}

\newcommand{\MYcomment}[1]{}

\newcommand{\MYnote}[1]{}

\newcounter{MYtablecntr}
\addtocounter{MYtablecntr}{1}

\newcommand{\MYlabel}{\small {$\bullet$}}

\newcounter{MYenumctrtwo}

\newcounter{MYenumctr}

\begin{document}

\title{A Case for Dataset Specific Profiling}

\author{
	$ {\textrm{Seth Ockerman}^{1,2} \textrm{, John Wu}^{2}\textrm{, Christopher Stewart}^{2}} $\\}
\affiliation{
     $\textrm{Grand Valley State University}^{1}\textrm{, The Ohio State University}^{2} $\\
     \country{}
     \institution{}
      \streetaddress{}
      \city{} 
      \state{} 
}




\begin{abstract}
Data-driven science is an emerging paradigm where scientific discoveries depend on the execution of computational AI models against rich, discipline-specific datasets.  With modern machine learning frameworks, anyone can develop and execute computational models that reveal concepts hidden in the data that could enable scientific applications. For important and widely used datasets, computing the performance of every computational model that can run against a dataset is cost prohibitive in terms of cloud resources.  Benchmarking approaches used in practice use representative datasets to infer performance without actually executing models.  While practicable, these approaches limit extensive dataset profiling to a few datasets and introduce bias that favors models suited for representative datasets. As a result, each dataset's unique characteristics are left unexplored and subpar models are selected based on inference from generalized datasets. This necessitates a new paradigm that introduces dataset profiling into the model selection process. To demonstrate the need for dataset-specific profiling, we answer two questions: (1) Can scientific datasets significantly permute the rank-order of computational models compared to widely used representative datasets?  (2) If so, could lightweight model execution improve benchmarking accuracy?  Taken together, the answers to these questions lay the foundation for a new dataset-aware benchmarking paradigm.
\end{abstract}



\maketitle

\section{Introduction}

With the rise of data-driven science, an increasing number of large, discipline-specific image datasets are being introduced to the public~\cite{defo-dataset,yang2020adaptive,yang2021uav,lan2021real}.  However, despite a widespread increase in datasets, an analysis by a team of Google and Berkeley researchers found that a very small number of datasets are used for computer vision benchmarking \cite{Koch_2021}. This is unsurprising given the popularity of measuring a learning architecture's efficacy by testing it against well-known datasets (e.g. Imagenet \cite{imagenet}, CIFAR-10 \cite{cifar}, MNIST \cite{mnist}, etc). The small number of benchmarking datasets has narrowed the scope of novel model design to a limited domain. 
Given the importance of data in the model design process and the relatively few datasets that are used for benchmarking, it is worth asking if they function well as representative datasets across different disciplinary areas.  In other words, are existing benchmarks good references for new datasets and corresponding workloads~\cite{stewart2008empirical,shen2009reference}. Furthermore, can we take for granted the unique influence that data has on all elements of the design process (CNN filter dimensions, hyperparameter searching, reaction rules, etc.) and assume that one learning architecture's success on one dataset will transfer to another? 

This study advocates for the adoption of the data-driven paradigm through individual dataset profiling. We will quantify the differences between widely used vision datasets used for benchmarking and discipline-specific datasets. We reexamine the use of representative datasets and prove that this practice can skew optimization away from each dataset's unique data paradigm. In addition to advocating for the adoption of data-driven design, we describe a potential solution that can address the cost barrier that drives the existing benchmarking paradigm. 

We recognize that individual dataset benchmarking is inherently difficult due to cost. For good reason, most researchers chose a single representative dataset to train and test their model against instead of running against every dataset in its application area. We describe early work that makes use of previous models' training cycles to predict a new model's final accuracy after just a few epochs of training, extending recent efforts to predict accuracy~\cite{Unterthiner_2020,Yamada_2016}.

This paper's main contributions are as follows. To prove the need for dataset-driven profiling, we profile a number of datasets across two model application areas: deep neural networks (DNNs) and particle swarm optimization (PSO) \cite{poli2007}. We examine the inherent differences in neural network performance on 8 different datasets through analysis of a variety of metrics. We also examine the differences in parameter estimations for 4 mass cytometry datasets. Finally, we present early work on an alternative method for DNN benchmarking that uses low-cost lightweight runs to predict test set performance after only a few epochs. 

\section{Related Work}

This section is organized as follows. Section \ref{sec:bias-in-benchmarking} will explore past work on dataset profiling and bias to prove the need for a new paradigm. Section~\ref{sec:surge-datasets} will describe new trends leading to more datasets and the challenges presented in classical benchmarking.  Section \ref{sec:sig-of-weights} will provide an overview of past work predicting neural network accuracy from weights. 

\subsection{Bias in Benchmarking Datasets}
\label{sec:bias-in-benchmarking}
It is standard design practice to test new models (trained weights, hyperparameter choices, and architecture itself) by running them against popular datasets like MNIST and CIFAR (\cite{cifar, mnist}). This is in part due to the popularity of dataset competitions as a method to popularize your learning architecture. However, this popularity comes with drawbacks. An analysis by \cite{everingham_2010} found no statistical difference between the performance of the top 10 algorithms in the 2010 PASCAL Visual Object Classes competition. This suggests the top algorithms are not fundamentally different from one another. Researchers worry that the lack of dataset diversity among popular datasets is causing models to learn from idiosyncrasies of the images rather than significant generalized characteristics (\cite{Ponce_2007, tor_2011}). \cite{tor_2011} also found that models trained on one representative dataset tend to test poorly on other representative datasets of the same category (i.e types of cars). This is not surprising given models tend to favor their own test sets. However, it is concerning that supposedly representative datasets do not create models with high enough levels of generalization to transfer to other similarly representative datasets. 

While the diversity of representative datasets has improved over time, they still suffer from limitations. Far more recent work (2021) found that neural networks were learning from noise in biomedical images datasets instead of the relevant medical features (\cite{Dhar_2021}).  These datasets were popular image benchmarks that many new algorithms and networks were tested against. While these datasets certainly provided a good sanity check for new approaches, they also can determine the success or failure of a new approach. This bias, as well as the similarity between the medical datasets,  can hamper novel approaches from seeing widespread adoption.  

Past works suggest current representative datasets have the potential to skew development toward a narrow solution space not representative of the complexity of real-world problems. In theory, the simple solution is to create a variety of datasets by sub-area which are perfectly representative. However, this is both impractical due to the black box nature of neural networks and impossible due to pure cost. A new benchmarking paradigm that enables lightweight low-cost model testing against specific datasets is needed. 

\subsection{A Surge in Datasets}
\label{sec:surge-datasets}
From astronomy~\cite{scalzo2010nearby,silverman2011fourteen} to agriculture~\cite{boubin2019autonomic,boubin2019managing,defo-dataset,zhang2020} to K-12 education~\cite{rao2018assessing,buele2020interactive} to software bugs and analysis~\cite{taylor2021study,taylor2021avis,stewart2013zoolander}, the velocity of dataset creation has surged in recent years.  While machine and deep learning algorithms seek to perform  classification and segmentation on all datasets, the semantics around these operations vary for each dataset. In astronomy, outlier pixels are associated with supernovae.  In agriculture, classification can distinguish healthy and unhealthy crops.  In education, clustering algorithms to identify students in need of additional tutoring.  We contend that these different use-cases impact the efficacy of machine learning algorithms, in terms of accuracy, training time and computational cost.  With the emergence of IoT for passive data collection, every field can create datasets to improve decision making and understanding.  

With the growing number of datasets, dataset management is increasingly important, especially for cost efficacy.  Data commons have emerged as storage repositories specifically for datasets~\cite{grossman2016case}.  In addition, research on cost effective replication~\cite{stewart2013zoolander,boubin2020softwarepilot,xu2016blending,du2022cost}, cloud processing models~\cite{dean2008mapreduce}, and other systems level techniques have matured. While we believe that these types of systems techniques also warrant dataset-specific data-driven approaches, in this paper, we are concerned with dataset efficacy rather than efficiency.  However, in future work, we will explore the unique computational 
demands imposed by different datasets.

\subsection{The Significance of Weights in Predicting Neural Network Accuracy}
\label{sec:sig-of-weights}
The primary way past works have attempted to predict the final accuracy of a network is through the use of early training curves (\cite{doman_2016}). However, both concurrent and recent work has shown a strong relationship between a network's weights and its characteristics and performance. \cite{Yamada_2016} was able to use weights obtained early in a neural network's training process to predict its eventual testing accuracy with higher accuracy than existing learning curve-based approaches. Very recently, \cite{Unterthiner_2020} found that using simple summary statistics based on network's fully trained weights could predict test set accuracy with an \(R^2\) score of over 0.98. This presents a compelling case for more investigation into the use of weights to predict neural network performance. 

 \cite{Yamada_2016} uses a variety of weight features to predict eventual accuracy based on early epoch weights. \cite{Unterthiner_2020} creates a dataset of 32,000 small-scale neural network's fully trained weights mapped to their final test set accuracy. They then use that dataset (dubbed CNN Zoo) to train gradient boosting machines (GBMs) to predict the test set accuracy of large-scale models.   While our early work (explored in \ref{sec:early_work}) is inspired by both \cite{Yamada_2016} and \cite{Unterthiner_2020}, it is also builds on them in a significant way. \cite{Unterthiner_2020} uses a small 4-layer network with randomly initialized weights. To build on this, we extract features from a large deep neural network that incorporates pre-trained ImageNet weights. We theorize that transfer learning will enable us to reduce the number of hyperparameter configurations needed to create a representative solution space (i.e scale of 1000s vs less than 100). \cite{Unterthiner_2020}'s work treated a network's training cycle as one unit of data to be mapped to final accuracy.  We instead record data at an epoch level. This enables the prediction of final accuracy based on only a few training epochs. In addition, both \cite{Yamada_2016} and \cite{Unterthiner_2020} focus on classical datasets. We instead focus on complex domain-specific datasets which better accounts for the potential bias of representative datsaets.

\section{Experimentation with Digital Agriculture and Deep Neural Networks}
\label{sec:dig_ag}
To prove the need for individual dataset profiling in the computer vision field, we compare model performance among domain-specific datasets (digital agriculture) and classical datasets. Our experiment will demonstrate two main ideas: 1) A set of models will rank differently relative to each depending on the category of dataset. 2) Each set of models ranks differently on an individual dataset basis. These ideas will demonstrate that representative datasets will always introduce some level of bias into the neural network design process. 

\subsection{Methodology}
\label{sec:distance-measure}
\begin{figure}[h]
    \centering
      \includegraphics[scale=0.40]{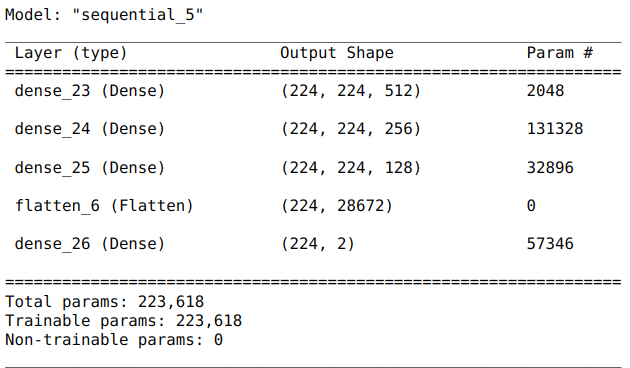}
    \caption{Small Fully Connected Classification Network}
    \label{fig:tag-on}
\end{figure}

To  measure the distance between model rankings among different datasets, we select four prevalent model architectures: InceptionV3, VGG16,  EfficentNet, ResNet50 (\cite{resnet, vgg16, InceptionV3, efficentNet}). Each network's weights are randomly initialized. A small fully connected classification network is placed on top of each network to allow for variation in the number of classes (a sample fully connected network is shown in figure \ref{fig:tag-on}). We also include an early stopping mechanism to prevent overfitting and provide standardization across different datasets. Our early stop callback will end training after the validation accuracy has stopped improving for 10 epochs.

We train and test these networks using two categories of datasets: classical datasets and digital agriculture datasets. We select 4 classical datasets based on popularity: CIFAR-10, CIFAR-100, imagenette2 (a subset of Imagenet), and MNIST (\cite{imagenet,cifar,mnist}). The digital agriculture datasets selected are as follows: fruits-360, PlantVillage, weed seedlings, and leaf defoliation dataset (\cite{weedSeedlings,plant-village,fruit-360,defo-dataset}). We select these datasets because they each represent a fundamental task in digital agriculture (e.g. fruit classification, drone-based defoliation detection, diseased plant classification, etc.). More details on each dataset can be found in table \ref{tab:dataset-comp}.

\begin{table} [h]
\centering
\begin{tabular}{| l | l | l |}
 \hline
 \textbf{Dataset} & \textbf{Classes} & \textbf{Images}  \\
\hline
CIFAR-10 & 10 & 60,000 \\
\hline
CIFAR-100 & 100 & 60,000 \\
\hline
MNIST & 10 & 60,000 \\
\hline
Imagenette & 10 & 13,000 \\
\hline 
Leaf Defoliation Dataset & 2 & 97,395 \\
\hline 
Fruits-360 & 131 &  90,483 \\
\hline 
PlantVillage & 38 & 87,000 \\
\hline
Weed Seedlings & 8 & 34,666 \\
\hline 
\end{tabular}
\caption{Individual Dataset Breakdown}
\label{tab:dataset-comp}
\end{table}

We run all four networks against the eight datasets and collect five metrics: test set accuracy at the 20\% training epoch, 75\% training epoch, and the last training epoch; sparse categorical crossentropy loss; and epochs run. In this case, the epoch metric functions as a loose cost approximation because the early stopping mechanism attempts to detect the number of epochs required for convergence.

Using these metrics, we create ranking vectors for each dataset. Each model is mapped to an arbitrary index in the vector, resulting in a base encoding vector as follows <InceptionV3, VGG16, EfficentNet, ResNet50>. For each metric and specific dataset (i.e accuracy, loss, or epochs run), we enter the ranking of that model into its encoded index. For example, the Leaf Defoliation ranked by accuracy would create the vector <3, 4, 1, 2>. In contrast, MNIST ranked by accurary would create the vector <1, 2, 3, 4>. We repeat this process for all eight datasets and create five ranking vectors per dataset.


By focusing on rankings, we isolate and standardize the effectiveness of a given DNN relative to a specific dataset. This avoids the inherent bias of comparing network accuracy across different datasets, which is often common practice. To measure the distance between rankings, we select CIFAR-10 as our base vector (\cite{cifar}). To measure the distance between datasets, we use  Euclidean and Kendal Tau distances.

\subsection{Results: The Inherent Differences of Classical and Discipline Specific Datasets}
\label{sec:results-distance}
After running tests across a variety of metrics, we find that digital agricultural datasets display fundamentally different properties than classical datasets. Sections \ref{sec:acc}, \ref{sec:loss}, \ref{sec:epoc} , and \ref{sec:rel-ld} will explore our results graphically. Section \ref{sec:num-a} will provide a brief numerical analysis of our results related to accuracy.

\subsubsection{Accuracy}
\label{sec:acc}
\begin{figure}[h]
    \centering
      \includegraphics[scale=0.25]{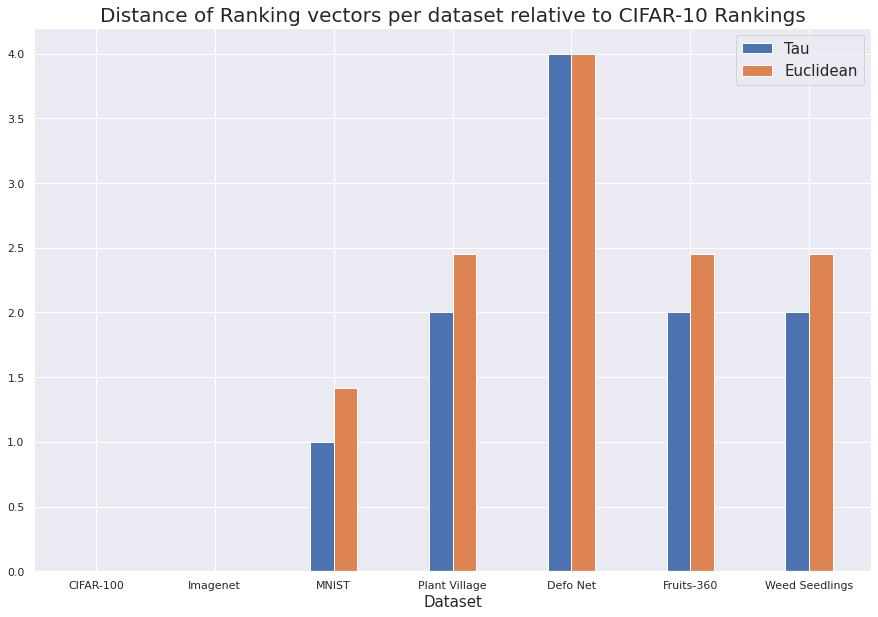}
    \caption{Final Test Set Accuracy Rankings relative to CIFAR-10}
    \label{fig:three-acc-metrics}
\end{figure}

Focusing specifically on final testing accuracy, we discover a significant difference between the rankings of digital agriculture and classical datasets. Figure \ref{fig:three-acc-metrics} shows two different distance metrics that measure the distance between a dataset's accurary ranking vector relative to CIFAR-10's accuracy ranking vector. Both CIFAR-100 and ImageNet have identical ranking vectors to CIFAR-10. In contrast, all four of the digital agriculture datasets create significantly different rank vectors. This indicates that for the purpose of testing models against other classical datasets, such as CIFAR-10, using a different classical dataset is an excellent benchmarking technique. However, against domain-specific datasets, such as digital agriculture datasets, the same models perform at highly different levels of effectiveness. These findings support the need for dataset aware benchmarking.  

\subsubsection{Loss}
\label{sec:loss}
In addition to final test set accuracy after training, each dataset shows different loss rankings. Figure \ref{fig:ale-cifar} shows the distance between a given dataset and CIFAR-10 based on three types of ranking vectors: final test set accuracy, loss, and epochs run. Once again, both CIFAR-100's and ImageNet's loss ranking vectors are identical to CIFAR-10. PlantVillage and the Leaf Defoliation dataset vectors are significantly different from all of the classical image datasets. Interestingly, MNIST and Fruits-360 display the same distance from CIFAR-10, while Weed Seedlings is identical to CIFAR-10 in terms of loss. We theorize this is due to the relative simplicity of Fruits-360 and Weed Seedlings compared to the other digital agriculture datasets. Pure object classification (fruit, weeds) is a simple task compared to detecting subtle differences that indicate defoliation or disease. This would allow for higher levels of generalization, enabling similar performance of standardized models from one dataset to the next. In the future, an investigation into an automatic dataset complexity analysis might be warranted to improve the dataset-aware benchmarking paradigm. Regardless, the discipline-specific datasets show an average higher distance in loss rankings compared to the classical datasets.

\subsubsection{Epochs}
\label{sec:epoc}
\begin{figure}[h]
    \centering
      \includegraphics[scale=0.20]{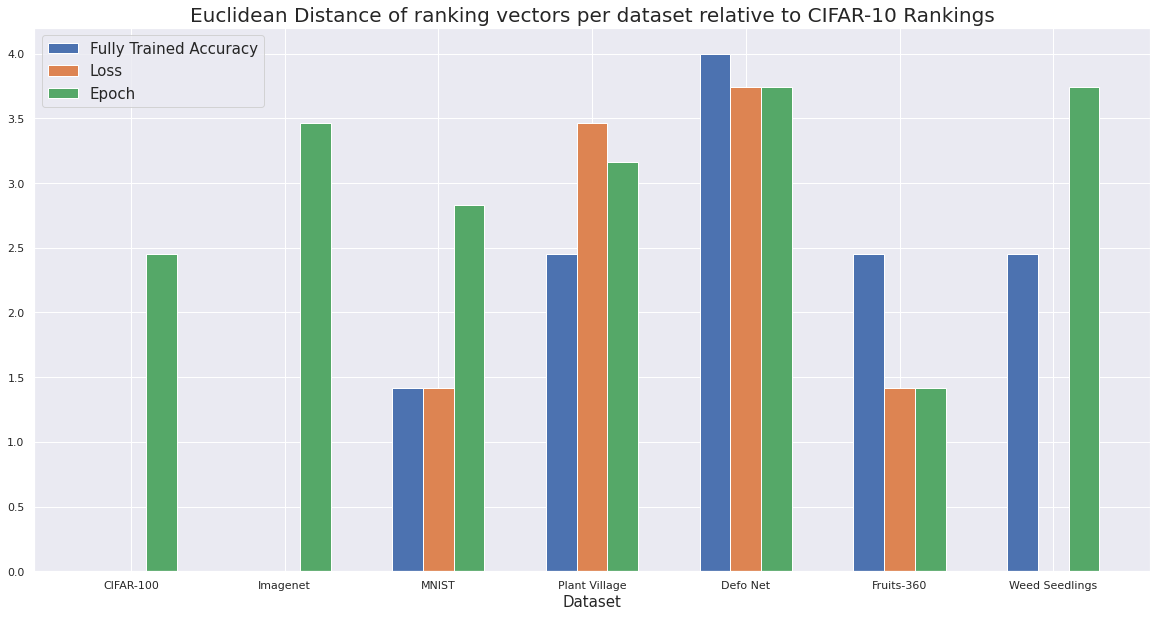}
    \caption{Euclidean Distance of Ranking vectors relative to CIFAR-10}
    \label{fig:ale-cifar}
\end{figure}

The green bar in figure \ref{fig:ale-cifar} represents distance between epoch ranking vectors relative to CIFAR-10. The Epoch ranking measured the number of epochs the model ran until the early stopping mechanism was triggered. This very loosely allows us to estimate the cost of training a given network. We use this to create a ranking vector of epochs (ranked in ascending order to prioritize low-cost networks). Our ranking system is by no means a perfect measure of cost because time per epoch can vary greatly between datasets,  but cost estimation is not the focus of this paper. We discover that per dataset there is a great degree of variance in ranking vectors.  Among ranking vectors, epoch vectors showed by far the most diversity compared to any of the other metrics we collected. The general grouping between classical and digital agricultural datasets that is shown in the accuracy and loss rankings is not shown in the epoch rankings. The lack of grouping by dataset category suggests that convergence significantly varies even within specific categories of datasets, which is unsurprising given convergence is notoriously difficult to estimate. The variance in epoch rankings by dataset supports the need for dataset aware benchmarking, particularly in the area of cost estimation. While our study focuses on estimating test set accuracy, a better method of cost estimation per dataset is a potentially interesting future study. 

\subsubsection{Relative to Leaf Defoliation Dataset}
\label{sec:rel-ld}
\begin{figure}[h]
    \centering
      \includegraphics[scale=0.20]{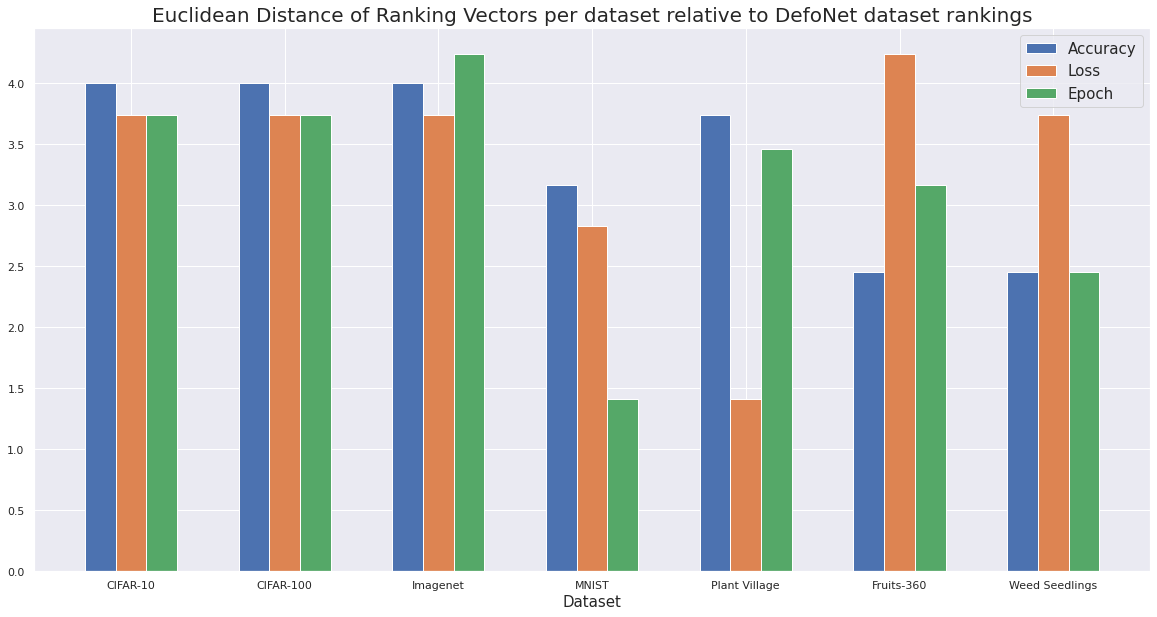}
    \caption{Euclidean Distance of Ranking vectors relative to Leaf Defoliation Dataset}
    \label{fig:ale-defonet}
\end{figure}

We discovered a strong grouping of classical dataset rankings, so we investigate if that same grouping exists among the three digital agriculture datasets we selected. To do this, we measure the distance between ranking vectors relative to the Leaf Defoliation Dataset instead of CIFAR-10. Figure \ref{fig:ale-defonet} shows the distance between final testing accuracy, loss, and epoch vectors relative to the Leaf Defoliation dataset. We clearly see the general grouping of three of the four classical datasets across all three metrics. MNIST is similar to the other classical datasets in terms of accuracy, but varies in loss and epochs. This same degree of consistency does not exist among the digital agricultural datasets. There are some similarities between Weed Seedlings and Fruit-360, however, PlantVillage remains distinct compared to its digital agricultural counterparts. These findings suggest more significant variation between the rankings of models on digital agricultural datasets and classical image datasets. 

\subsubsection{Numerical Analysis}
\label{sec:num-a}

To try to improve our insights into the difference between accuracy rankings across different datasets, we calculate each datasets accurary vector distance relative to all other datasets. We repeat this process and create average distances from each category of dataset. The average distance of classical datasets from itself is 0.71, while the average distance of a digital agriculture dataset from a classical dataset is 2.67. The almost four times increase in ranking distance shows an inherent difference in model performance by dataset group, suggesting that testing digital agriculture models on classical datasets will bias them away from their own data paradigm.

 The average distance of digital agriculture datasets from themselves is 2.69, while the average classical datasets distance from digital agriculture datasets is 2.67. The near equivalent distances suggest that model performance varies significantly at a dataset level even among groupings such as digital agriculture. The increased variation in model performance among digital agriculture datasets can partially be explained by the large discrepancy in distance between self accuracy of classical datasets (0.71) and digital agriculture datasets (2.69). These findings indicate that classical datasets are significantly better at benchmarking themselves than digital agriculture datasets are. We theorize the increased difficulty in self-benchmarking is due to the high diversity in discipline-specific datasets when compared to classical datasets. Generalized groupings such as digital agriculture datasets or classical datasets are not adequate to benchmark niche and domain-specific applications. A new paradigm for benchmarking is needed. A summary of all numbers discussed in this section can be found in tables \ref{tab:dataset-avg-dist} and \ref{tab:dataset-group-dist}.

\begin{table} [h]
\centering
\begin{tabular}{| l | l | l |}
 \hline
 \textbf{Dataset} & \textbf{Classical} & \textbf{Argo.}  \\
\hline
CIFAR-10 & 0.47 & 2.84 \\
\hline
CIFAR-100 & 0.47 & 2.84 \\
\hline
MNIST & 1.41 & 2.20 \\
\hline
Imagenette & 0.47 & 2.84 \\
\hline 
Leaf Defoliation Dataset & 3.79 & 2.88 \\
\hline 
Fruits-360 & 2.19 &  2.06 \\
\hline 
PlantVillage & 2.54 & 3.74 \\
\hline
Weed Seedlings & 2.19 & 2.06 \\
\hline 
\end{tabular}
\caption{Average distance of a dataset from datasets of its type}
\label{tab:dataset-avg-dist}
\end{table}

\begin{table} [h]
\centering
\begin{tabular}{| l | l | l |}
 \hline
 \textbf{} & \textbf{Classical} & \textbf{Argo.}  \\
\hline
\textbf{Classical} & 0.71 & 2.68 \\
\hline
\textbf{Argo.} & 2.68 & 2.69 \\
\hline
\end{tabular}
\caption{Average dataset category distance}
\label{tab:dataset-group-dist}
\end{table}

\subsection{Summary}
We demonstrated that neural networks perform differently on classical datasets than on domain-specific areas datasets such as digital agriculture. The average distance of rankings between a digital agriculture dataset and a classical dataset is nearly 4 times the distance of a classical dataset from itself. Additionally, we discover that digital agriculture datasets are not significantly better for benchmarking themselves than classical datasets. As such, the arbitrary grouping of dataset by sub-area (i.e. classical vs digital agriculture) is not a solution to address differences between datasets. The inherent difference between complex datasets makes it nearly impossible to perform nonbiased benchmarking through the use of a single representative dataset. Domain-specific datasets are not sufficiently better at benchmarking than classical datasets. All of this together demonstrates the need for the adoption of individual dataset profiling in the neural network community.

\section{Particle Swarm Optimization in Mass Cytometry Datasets}
\label{sec:PSO}
This difference in datasets is further seen in even niche domains outside of vision and neural network learning architectures. Such is the case in Mass Cytometry datasets where biophysicists commonly model biological systems using ordinary differential equation (ODE) reaction networks \cite{rohrs2019,loskot2019,la2018}. In these networks, parameter estimation, specifically of rate constants, is performed with a variety of "learning" or optimization heuristics. One commonly used heuristic is the biologically inspired algorithm, particle swarm optimization (PSO). In this heuristic, particles search the parameter-cost space for the optimal set of rate constants that best fit the observed trajectories of time-stamped abundance data in Mass Cytometry. The PSO's update mechanism we apply in our methodology is driven by three weights, (1) the influence of a particle's current best estimate, (2) the influence of the global best estimate, and (3) each particle's inertia. To show the aforementioned differences in learning architecture performance by dataset, we provide euclidean distances of ranking vectors of five configurations of PSO weights for four different mass cytometry datasets.
\subsection{Methodology}
The four datasets contain both experimental and simulated data. In the case of real data, time-stamped protein species of CD8 T cells \cite{CD8} and CD56 cells \cite{NKG2D} were measured. For the two simulated datasets, initial conditions were randomly sampled from multivariate lognormal distributions. Then, using the ODE reaction networks defined in Bionetgen \cite{bionetgen}, initial conditions were evolved to set time points. Each dataset's time points are shown in Table \ref{tab:Dataset Times}.
\begin{table} [h!]
\centering
\begin{tabular}{ |l|l| } 
 \hline
 Dataset & Times (minutes) \\
 \hline
 CD56 NK Cells & 16, 32 \\ 
  \hline
 CD8 T Cells & 1, 2\\ 
  \hline
 Simulated 1 & 0, 0.5, 2, 10, 20, 30  \\ 
  \hline
 Simulated 2 & 0, 1.5\\ 
 \hline
\end{tabular}%
\caption{\label{tab:Dataset Times} Times Points in Each Dataset}
\end{table}

\begin{figure} [h]
     \centering
     \includegraphics[width=15cm,height=6cm,keepaspectratio,]{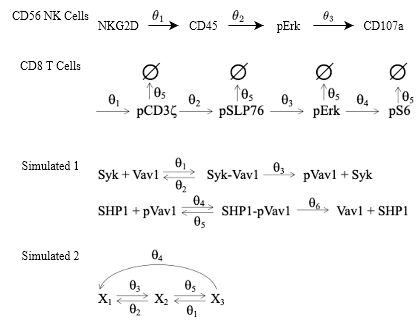}
     \caption{ODE Reaction Networks used in Bionetgen}
     \label{fig:schematic}
\end{figure}

All reaction networks are shown and defined in Figure \ref{fig:schematic}. Note that the $\theta_i $'s are the parameters being estimated and that the labels on the left correspond to their respective datasets. Each of the five different configurations of PSO weights and their respective labels are shown in Table 5. To provide reasonable computational constraints in the context of run time costs, each PSO configuration was standardized to 200 particles and 20 epochs. Each PSO configuration is run 30 times against each dataset, giving us a set of estimates for each dataset's respective ODE models. A more rigorous explanation of the PSO used to generate the data can be found in \cite{GMM}. 

\begin{table} [h]
\centering
\resizebox{\columnwidth}{!}{%
\begin{tabular}{ |l|l|l|l| } 
 \hline
 Configuration & Particle Best Weight & Global Best Weight & Particle Inertia \\
 \hline
 A & 3.0 & 1.0 & 6.0\\ 
  \hline
 B & 4.0 & 2.0 & 5.0\\ 
  \hline
 C & 5.0 & 3.0 & 4.0\\ 
  \hline
 D & 5.0 & 2.0 & 10.0\\ 
  \hline
 E & 3.0 & 4.0 & 3.0\\ 
 \hline
\end{tabular}%
}
\caption{\label{tab:PSO Configs} PSO Weight Configurations}
\end{table}
We ranked each PSO configuration by the average standard deviation of estimates and by cost. In this case, smaller deviations and costs are ranked higher (i.e the smallest standard deviation PSO configuration would be ranked one). We define cost to be the square difference of means, variances, and covariances between the observed data and the data generated from estimates. We chose these two metrics because the range of estimates roughly indicates the efficiency of PSO estimation while the cost indicates a level of dataset fit.

Similar to the ranking method performed in Section 3, we encode configurations of PSO to a vector. In this case, we map each configuration to an index in the vector, creating the mapping <A, B, C, D, E> (e.g. Configuration A is mapped to index 0 of the vector). Using this mapping, ranking vectors of PSO configurations were formed for each dataset. For instance, when estimating using the CD8 T Cells dataset, a ranking of D, E, C, B, and A in descending order produces the vector <5, 4, 3, 1, 2>). Once encoded, relative distances of ranking vectors were computed with respect to each dataset.

\subsection{Differences Across Mass Cytometry Datasets}
\begin{figure} [h]
     \centering
     \includegraphics[width=15cm,height=6cm,keepaspectratio,]{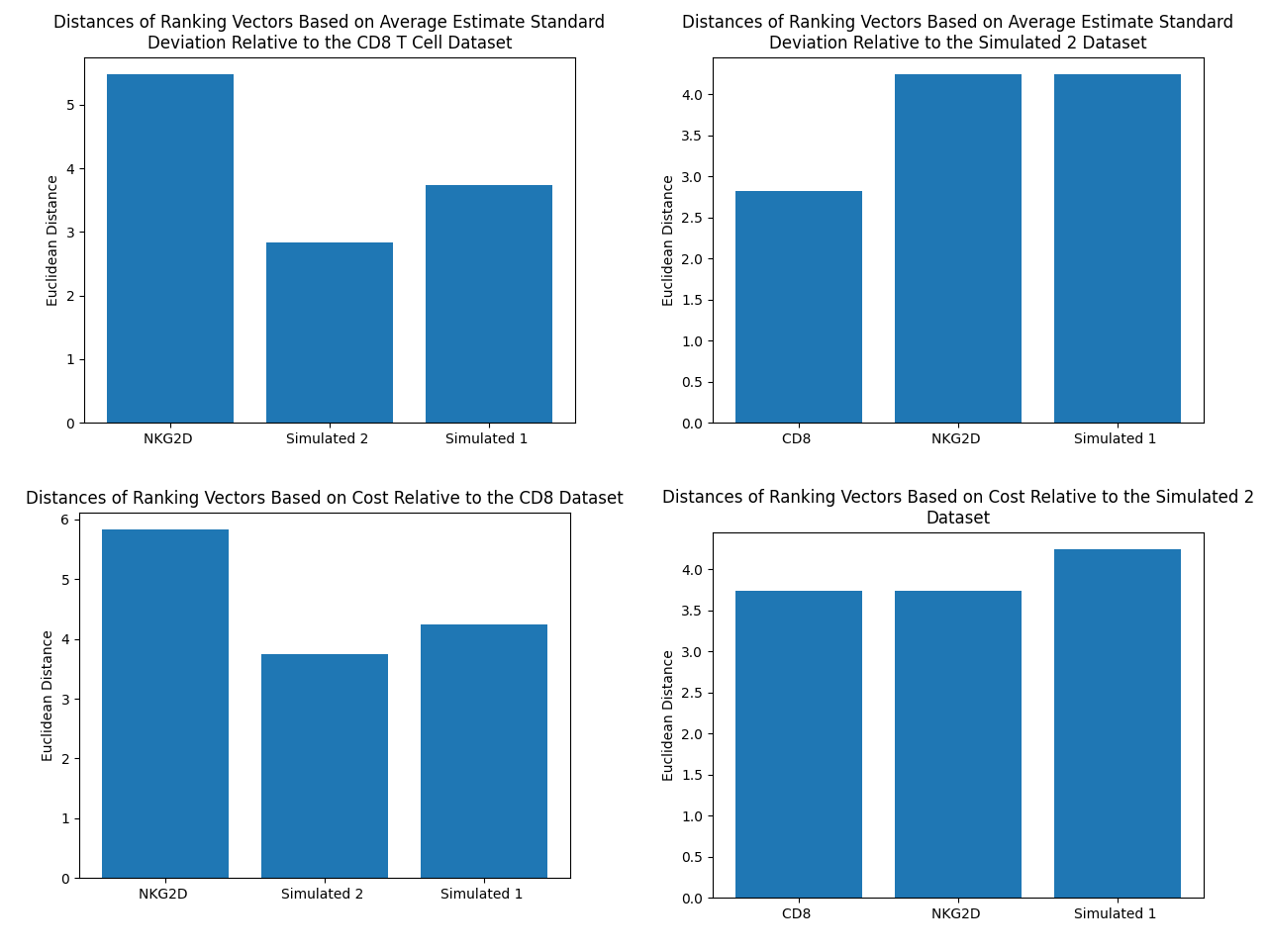}
     \caption{Distances Between Ranking Vectors of PSO Configurations}
     \label{fig:psoDist}
\end{figure}
For sake of brevity, only the euclidean distances with respect to the CD8 T cells and Simulated 2 are shown in Figure \ref{fig:psoDist}. As shown by the bottom two bar charts in Figure \ref{fig:psoDist}, in terms of cost, rankings dramatically differ. This difference in rankings implies that the optimal PSO configuration in terms of dataset fit can be dataset-specific. 

Furthermore, this trend continues to be seen in the standard deviation rankings. In this case, smaller standard deviations imply that each repeated PSO execution converges on a smaller interval of estimates.   Figure \ref{fig:psoDist} reveals that certain PSO configurations converge faster depending on the dataset.

Also, observe that the ranking differences remain between the two synthetic datasets where although reaction networks differ their source probability distributions remained the same, demonstrating that structural differences within a dataset may lead to different optimal PSO configurations. 

Even in fields outside of computer vision such as computational biology, datasets fundamentally influence model performance. Ultimately, these discrepancies further reinforce the need for dataset specific profiling and benchmarking. 
\section{Early Work on Lightweight Model Profiling}
\label{sec:early_work}
Sections \ref{sec:dig_ag} and \ref{sec:PSO} demonstrated that reusing learning architectures between datasets reduces the overall effectiveness of each dataset-model combination. A simple but effective solution to the demonstrated problem is to view the dataset profiling process as part of hyperparameter searching. However, profiling each dataset-model combination makes the naive assumption that cost can grow towards infinity. Individual dataset profiling is not feasible on a large scale for the majority of developers. A new method of model benchmarking and dataset profiling is needed. This section details our experimentation using early training weights to predict final training accuracy.

\subsection{Methodology}
\label{sec:setup-damb}
We design an experiment to test the use of weights to predict final testing accuracy on complex a domain-specific dataset: the Leaf Defoliation Dataset (\cite{defo-dataset}.  To create a solution space, we test a variety of hyperparameter configurations using VGG16 DNN with pre-trained ImageNet weights combined with a small, fully connected classification neural network (a sample is shown in \ref{fig:tag-on}). 

We chose 35 different hyperparameter configurations to explore that vary optimizer, learning rate, and final layer activation function. We run each of the 35 configurations for 75 epochs and record the final test set accuracy of each configuration.

Each epoch we save the weights to extract summary statistics from them. We calculate the mean, variance, and q-th percentiles where \(q \in \{0,25,50,75,100\}\) (\cite{Unterthiner_2020}) for biases and kernel weights separately. WWe calculate these statistics for each neural network layer, creating a 2x7 vector for each layer. Combining all 17 layers into a single matrix, we generate a 17x2x7 representation of means, variances, and percentiles. This matrix is then mapped to the final testing accuracy of its respective model's configuration. Because we create mappings at an epoch level, we significantly increase the sample space we explore. In total, we create 2625 accuracy mappings from the Leaf Defoliation Dataset. 

We perform an 80/20 train/test split of our vector accuracy mappings. In contrast to typical train/test splits, we do not randomize the placement of the mappings. Instead, we ensure the 20\% in the test set is composed entirely of hyperparameter combinations that do not exist in the training set. The nonrandom nature of the test set is to prevent overfitting and leakage from the training dataset to the test set. Our approach results in a training set with 31 hyperparameter configurations and a testing set with four unseen hyperparameter configurations. 

For prediction, we select gradient boosting machines implemented in XGBoost's gradient boosting forest package (\cite{Chen_2016}).  We split our training set into a train and validation set at an 80\% 20\% ratio. Using that validation set, we perform hyperparameter tuning, resulting in a model of 128 estimators with a max depth of 7 per tree. A full breakdown of all selected hyperparameters can be found in \ref{tab:gbm-hyper}.

\begin{table} [h]
\centering
\begin{tabular}{| l | l |}
 \hline
 \textbf{Hyperparameter } & \textbf{Value}  \\
\hline
Objective & Regressive: Linear \\
\hline
Column Sample By Tree & 0.3 \\
\hline
Learning Rate & 0.1 \\
\hline
Max Depth & 7 \\
\hline
 Alpha & 40 \\
\hline
Number of Estimators & 128  \\
\hline
\end{tabular}
\caption{GBM hyperparamater selection}
\label{tab:gbm-hyper}
\end{table}

The testing data consists of the same number of hyperparameter configurations each time. However, we vary the percent of each model configuration's training cycle we include in the prediction process.  By limiting the data inputted to the GBM to an artificial n-th epoch of training time, we simulate lightweight benchmarking runs. For example, by reducing input data to the first five epochs of weight data, we test the accuracy of predictions given only a fraction of the total training time. We select epochs 4, 8, 19, 38, 56, 76 which represent 5\%, 10\%, 25\%, 50\%, 75\%, and 100\% of the training time respectively.

\subsection{Results}
After hyperparameter tuning on the validation set, we test our trained model on each of the test splits. Using the entirety of the test data, we achieve an accuracy of 81.33\% and a relative root mean squared error (RRMSE)  of 0.196. Crucially, there is a minimal decrease in accuracy if we reduce the number of epochs we use as an input for our test set. Using only 4 epochs of input data from the test set (representing roughly 5\% of its training time) we achieve an accuracy of 80.61\% and an RRMSE of 0.201. This trend continues across all designated input data splits. Across all splits, there is less than a 1\% change in accuracy and RRMSE. This indicates that the number of epochs of input data has little impact on the accuracy of a fully trained model. A full breakdown of accuracy and RRMSE by epoch of input data can be found in \ref{tab:vary_input_results}.

\begin{table} [h]
\centering
\begin{tabular}{| l | l | l | l|}
 \hline
 \textbf{Epoch } & \textbf{Training Time (\%)}&\textbf{Accuracy} & \textbf{RRMSE} \\
\hline
4 & 5\% & 80.61\% & 0.201 \\
\hline
8 & 10\% & 80.84\% & 0.199 \\
\hline
19 & 25\% & 81.09\% & 0.197 \\
\hline
38 & 50\% & 81.24\% & 0.197 \\
\hline
56 & 75\% & 81.31\% & 0.196 \\
\hline
75 & 100\% & 81.33\% & 0.197 \\
\hline
\end{tabular}
\caption{Accuracy and RRMSE of different percents of test set model training time}
\label{tab:vary_input_results}
\end{table}

\section{Conclusion}
In section \ref{sec:dig_ag}, we demonstrated that discipline-specific datasets significantly alter model performance compared to widespread classical benchmarking datasets. In addition, we discovered that the complexity of discipline-specific datasets causes them to suffer similar benchmarking limitations to classical datasets.

In section \ref{sec:PSO}, we explored three key findings: (1) that dataset differences in "learning" is not limited to neural networks, (2) that this discrepancy remains within datasets of even a highly specific domain such as computational biology, and (3) that this difference in ranking vectors applies to both experimental as well as synthetic data where a dataset's probability distribution is known.

Our findings demonstrate that datasets shape model performance in fundamental ways, necessitating that dataset profiling becomes part of the machine learning design process. 

\subsection{Vision for the Future Of Our Early Work}
Using GBMs, we achieve an accuracy of roughly 81\% across all different splits of input data. Our early work (described in section \ref{sec:early_work}) has not progressed enough to significantly change benchmarking practices. We hope, however, that our work will start a discussion about what a more developed dataset-specific benchmarking system could look like.  It is especially promising that changing the percent of training time that was inputted into our model had little to no effect on its overall accuracy. Given a model trained on robust data, predicting the final accuracy of a neural network is possible after only a small number of epochs. This finding transforms lightweight weight-based benchmarking from a niche theory into a plausible widespread reality. However, for this to be possible,  we need to save the weights of networks as they are trained and tested against a dataset. Using this approach, we could construct a weight-based solution space per dataset. This solution space could then be used to predict the performance of future learning architectures after just a few training epochs. Early prediction of model accuracy would exponentially reduce the cost of testing new architectures against new datasets, enabling more domain-specific models to emerge. 

Despite being limited in time and resources,  our approach showed initial success in predicting final network accuracy after only a few epochs. In future work, we hope to build on our findings by testing other types of neural networks (e.g. using a DNN instead of PSO on mass cytometry) and data-driven changes (explore more hyperparameter configurations, more epochs, more models, etc). We also hope to experiment with transferable prediction mechanisms to create a unified solution space for vastly different model architectures. 
{\footnotesize
\bibliographystyle{abbrv}
\bibliography{references}
}

\end{document}